%% file: main.tex
\documentclass[10pt,twocolumn,letterpaper]{article}

\usepackage{iccv}              % To produce the CAMERA-READY version
\usepackage{algorithm}
\usepackage{algorithmic}
\usepackage{amsmath}

\usepackage{amssymb}
\usepackage{bbding}
\usepackage{pifont}
\usepackage{xcolor}
\usepackage{tikz}
\usepackage[accsupp]{axessibility}  % Improves PDF readability for those with disabilities.
\definecolor{cjcolor}{RGB}{0,0,255}
\newcommand{\cj}[1]{{\color{cjcolor}#1}}


\definecolor{iccvblue}{rgb}{0.21,0.49,0.74}
\usepackage[pagebackref,breaklinks,colorlinks,allcolors=iccvblue]{hyperref}

\newcommand\blfootnote[1]{%
\begingroup
\renewcommand\thefootnote{}\footnote{#1}%
\addtocounter{footnote}{-1}%
\endgroup
}

\def\paperID{ } % *** Enter the Paper ID here
\def\confName{ICCV}
\def\confYear{2025}

\title{\textit{RoboTron-Nav}: A Unified Framework for Embodied Navigation Integrating Perception, Planning, and Prediction}

\author{Yufeng Zhong$^{\dagger}$ \quad Chengjian Feng$^{\dagger}$ \quad Feng Yan \quad Fanfan Liu \quad Liming Zheng \quad Lin Ma$^{\ddagger}$ \\ \\
{Meituan}\\
\url{https://yvfengzhong.github.io/RoboTron-Nav}
}
\begin{document}
\maketitle

\blfootnote{$\dagger$ Equal contribution. $\ddagger$ Corresponding author.}

\input{sec/0_abstract}    
\input{sec/1_intro}
\input{sec/2_related_work}
% \input{sec/2-5_dataset}
\input{sec/3_method}
\input{sec/4_experiments}
\input{sec/5_conclusion}
{ 
\small 
\bibliographystyle{ieeenat_fullname}
\bibliography{main}
}

% WARNING: do not forget to delete the supplementary pages from your submission 
\input{sec/X_suppl}

\end{document}

%% file: sec/0_abstract.tex
\begin{abstract}
In language-guided visual navigation, agents locate target objects in unseen environments using natural language instructions. For reliable navigation in unfamiliar scenes, agents should possess strong perception, planning, and prediction capabilities. Additionally, when agents revisit previously explored areas during long-term navigation, they may retain irrelevant and redundant historical perceptions, leading to suboptimal results. In this work, we propose \textbf{RoboTron-Nav}, a unified framework that integrates {p}erception, {p}lanning, and {p}rediction capabilities through multitask collaborations on navigation and embodied question answering tasks, thereby enhancing navigation performances. Furthermore, RoboTron-Nav employs an adaptive 3D-aware history sampling strategy to effectively and efficiently utilize historical observations. By leveraging large language model, RoboTron-Nav comprehends diverse commands and complex visual scenes, resulting in appropriate navigation actions. RoboTron-Nav achieves an 81.1\% success rate in object goal navigation on the $\mathrm{CHORES}$-$\mathbb{S}$ benchmark, setting a new state-of-the-art performance. 
%Project page: \url{https://yvfengzhong.github.io/RoboTron-Nav/}.
\end{abstract}
% todo: 修改sota结果

%% file: sec/1_intro.tex
\vspace{-3mm}

\section{Introduction}
\label{sec:intro}
% \vspace{-1mm}
Embodied navigation~\cite{anderson2018evaluation} is a key component of embodied artificial intelligence~\cite{huang2025dadu, zheng2025dataplatter, yan2024robomm, zheng2024robocas, liu2024robouniview,luo2024autom3l}, especially language-guided visual navigation~\cite{krantz2020beyond, ku2020room, savva2019habitat, shen2019situational, campari2020exploiting, gupta2017cognitive, chaplot2020object, rudra2023contextual}. Unlike autonomous driving~\cite{huang2024drivemm}, language-guided visual navigation requires agents to comprehend natural language instructions and autonomously explore unseen visual environments to locate target objects, which presents unique challenges. Specifically, to reliably navigate in unfamiliar visual environments, agents must effectively perceive surrounding scenes, plan strategies to achieve their goals, and predict appropriate navigation actions.

% top出自spoc的high_quality的task=ObjectNavType,house=12375,sub_house_id=60_search-for-a-bowl
\begin{figure}[!htbp]
\centering
\includegraphics[width=0.47\textwidth]{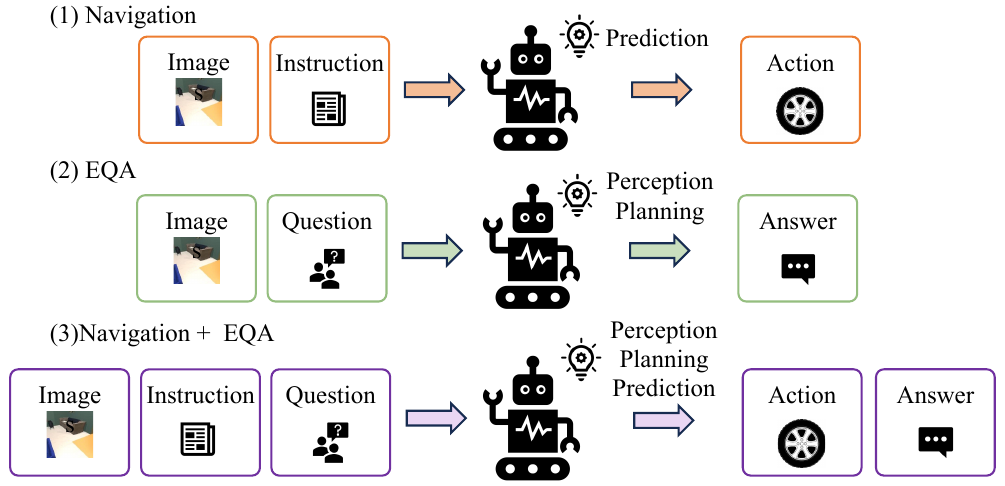}
\vspace{-2mm}
\caption{{Multitask collaboration.} In (1) navigation tasks, only actions are produced, missing the perception and planning found in (2) EQA tasks. (3) Multitask collaboration unifies perception, planning, and prediction for a more comprehensive model.}
\vspace{-4mm}
\label{fig:motivation}
\end{figure}

Recently, significant advancements in language-guided visual navigation have been driven by the robust understanding and generalization capabilities of visual-language models (VLMs)
% ~\cite{bordes2024introduction} 
that seamlessly integrate visual and textual information.
For example, in Object Goal Navigation (ObjectNav)~\cite{savva2019habitat, shen2019situational, campari2020exploiting, xie2023implicit, zhang2023layout, du2023object, gupta2017cognitive, chaplot2020object, rudra2023contextual, campari2022online, chang2020semantic, zhu2022navigating}, numerous approaches~\cite{xie2023implicit, zhang2023layout, du2023object, campari2022online, chang2020semantic, zhu2022navigating} have achieved impressive results by using VLMs to align and integrate textual and visual inputs, enabling agents to better understand complex environments and navigate to target objects more effectively.
Although current navigation models excel at locating target objects, they often struggle to provide efficient path planning and explain the reasoning behind their path choices, as shown in the \textit{first row} of Fig.~\ref{fig:motivation}.
This challenge is largely due to traditional navigation datasets focusing primarily on recording coordinate trajectories and neglecting the hierarchical reasoning process employed by humans. For instance, humans typically first identify rooms, then locate furniture, and finally search for specific objects.

\begin{figure}[!tbp]
\centering
\includegraphics[width=0.47\textwidth]{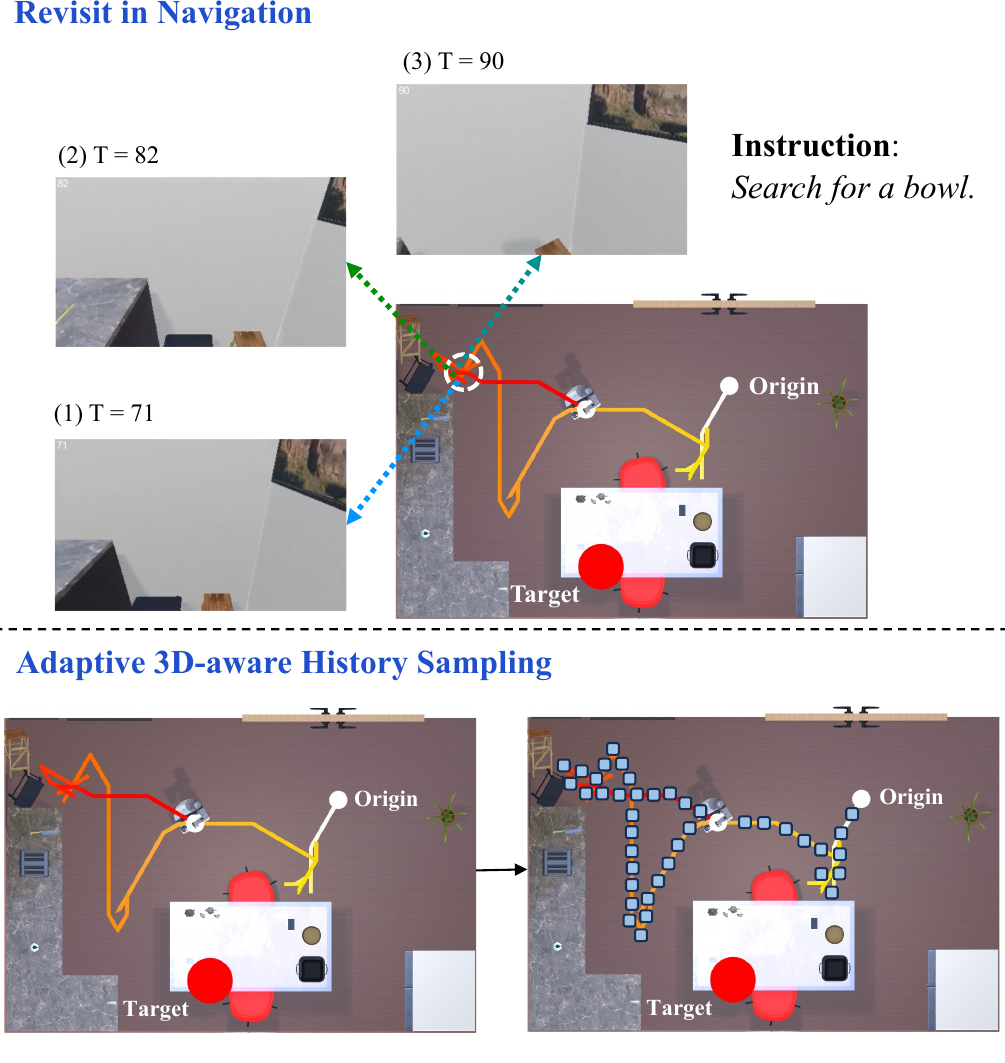}
\vspace{-4mm}
\caption{\textit{Top:} During long-term navigation, agents may revisit the same areas multiple times, making previous observations redundant. For instance, at $T = 71$, $T = 82$, and $T = 90$, the scenes observed are very similar, highlighting the redundancy in the agent's perception. \textit{Bottom:} By applying the {Adaptive 3D-aware History Sampling} strategy, we sample the dense and disorganized trajectory into sparse and organized steps, reducing observational redundancy while preserving visual information at key locations. The trajectory is produced by the SPOC-driven agent~\cite{ehsani2024spoc}.}
\vspace{-4mm}
\label{fig:path}
\end{figure}

Additionally, we have observed a significant amount of redundancy in observations during navigation, as agents frequently revisit the same locations multiple times. 
% When agents revisit previously explored areas during exploratory missions, they often retain irrelevant and redundant historical perceptions, leading to suboptimal prediction results. 
As illustrated at the \textit{top} of Fig.~\ref{fig:path}, the white circle marks areas that the agent revisits multiple times. At (1) $T = 71$ and (2) $T = 82$, the agent focuses on the countertop area and two chairs. At (3) $T = 90$, the scene remains largely unchanged, capturing an environment similar to that at (1) and (2) once again. These observations demonstrate that the scenes within the revisit areas are highly similar, indicating the redundancy present in the agent's perception. Therefore, when agents revisit the same area, previous observations may become redundant and should be discarded.

\textcolor{black}{This scenario raises two key questions driving our research: \textbf{(1) How can we clarify agents' decision-making processes? (2) How can we enhance long-term navigation efficiency?}}
To tackle these challenges, we propose {RoboTron-Nav}, a unified framework for embodied navigation that integrates {p}erception, {p}lanning, and {p}rediction. As illustrated in Fig.~\ref{fig:motivation}, RoboTron-Nav adopts a {Multitask Collaboration} strategy, leveraging joint training on both navigation and embodied question answering (EQA) tasks to fully harness perception and planning for improved prediction capabilities. \textcolor{black}{Within this framework, the EQA task seamlessly integrates perception and planning by requiring agents to answer questions based on their historical observations and navigational trajectories. To facilitate this, we construct EQA datasets that \emph{explicitly model the decision-making processes} involved in navigation, empowering agents to engage in deep reasoning and task planning akin to human cognition, and ultimately enhancing navigation interpretability. More details about our novel EQA dataset construction method are provided in Sec.~\ref{sec:dataset}.}

Furthermore, RoboTron-Nav employs an {Adaptive 3D-aware History Sampling} strategy to effectively and efficiently minimize observation redundancy by controlling the density and diversity of historical sampling in both spatial and semantic domains. 
% Specifically, this method selects RGB frames that are neither overlapping nor adjacent in spatial positions, or are captured from different viewpoints at the same location, to serve as valid observations.
{Specifically, we consider the RGB frames as valid observations only if they are either spatially non-adjacent to other frames, or spatially adjacent but captured from different viewpoints.} In addition, RoboTron-Nav introduces \emph{position-enhanced historical features} that utilize the agent's positions to augment historical semantic features with trajectory information, thereby preventing redundant exploration of the same location.
As illustrated at the \textit{bottom} of Fig.~\ref{fig:path}, \textcolor{black}{we sample the dense and disorganized trajectory to obtain sparse and organized steps}, minimizing redundancy while preserving crucial observations at key locations.
Finally, by leveraging the large language model (LLM), RoboTron-Nav can understand diverse instructions and complex visual scenes, resulting in effective navigation.
Through extensive experimentation, RoboTron-Nav has demonstrated remarkable capabilities, achieving an impressive 81.1\% success rate (SR) in ObjectNav on the $\mathrm{CHORES}$-$\mathbb{S}$ benchmark~\cite{ehsani2024spoc}. This performance not only sets a new state-of-the-art but also represents a substantial 9\% absolute improvement over previous methods.

% The main contributions of this paper include: 
To summarize, we make the following contributions:
% (1) We introduce \textbf{RoboTron-Nav}, a unified framework integrating \textbf{P}erception, \textbf{P}lanning, and \textbf{P}rediction, enhancing navigation by {multitask collaboration} on navigation and EQA tasks. (2) RoboTron-Nav employs the {adaptive 3D-aware history sampling} strategy, effectively utilizing historical observations by selecting non-overlapping RGB frames to reduce redundancy. (3) RoboTron-Nav achieves an 75\% success rate in ObjectNav on the $\mathrm{CHORES}$-$\mathbb{S}$ benchmark, setting a new state-of-the-art, with a 18\% absolute improvement over previous methods.
\begin{itemize}
    \item We introduce {RoboTron-Nav}, a unified framework integrating {p}erception, {p}lanning, and {p}rediction, enhancing navigation by {multitask collaboration} on navigation and EQA tasks.
    \item RoboTron-Nav utilizes an {adaptive 3D-aware history sampling} strategy, which minimizes observational redundancy by regulating the density and diversity of historical sampling across both spatial and semantic domains.
    \item RoboTron-Nav achieves an {81.1\%} SR in ObjectNav on the $\mathrm{CHORES}$-$\mathbb{S}$ benchmark, establishing a new state-of-the-art, with a {9\%} absolute improvement over previous methods.
\end{itemize}

% \vspace{-2mm}

% RoboTron-Nav employs an adaptive 3D-aware history sampling strategy that leverages spatial location information to effectively reduce redundant observations, pioneering a novel position-based history sampling method in embodied navigation.

\begin{figure*}[!htbp]
\centering
\includegraphics[width=\textwidth]{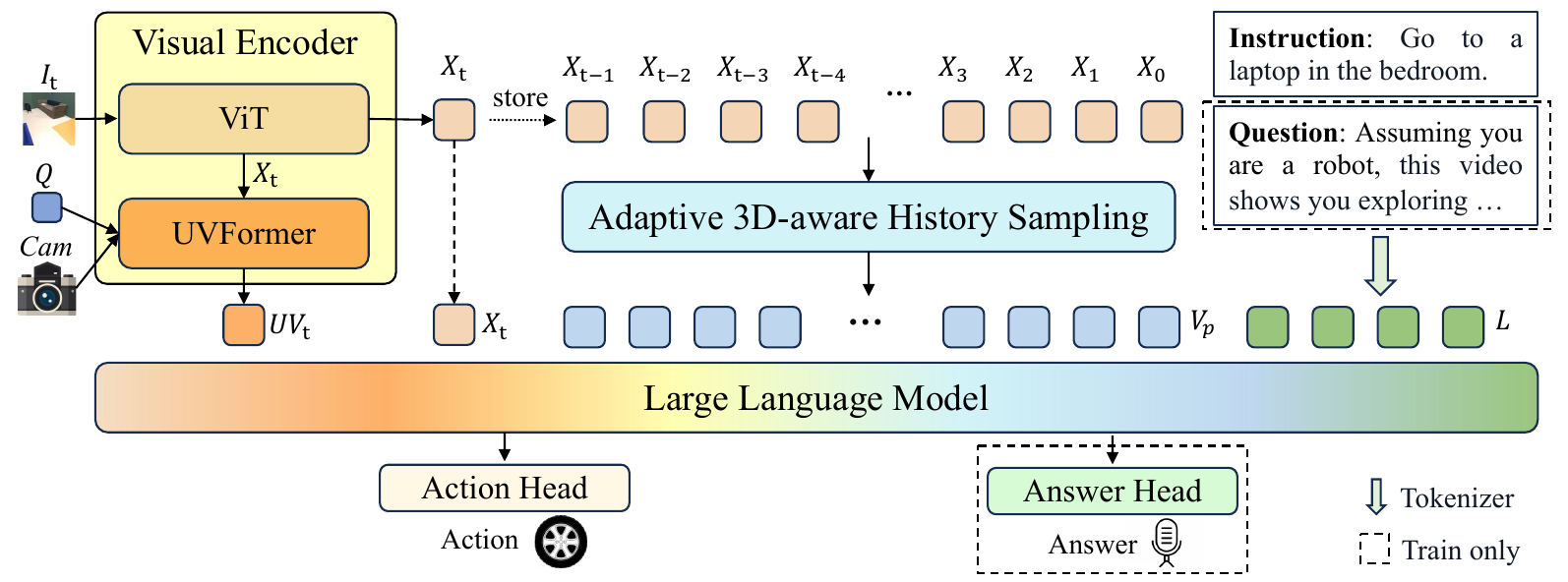}
% todo：caption中应该将train only和MC结合起来说明
\vspace{-8mm}
\caption{{Overview of RoboTron-Nav architecture.} The current frame $I_{t}$ is initially processed through 2D and 3D feature extraction using the visual encoder. Then, historical features are filtered through the {Adaptive 3D-aware History Sampling} strategy. The visual features obtained, along with the accompanying linguistic instructions, are then fed into the large language model (LLM). Leveraging the {Multitask Collaboration} strategy, which significantly enhances navigation capabilities through joint training on both navigation and EQA tasks, the LLM produces two outputs via multimodal fusion: executable navigation actions and natural language answers.}
\vspace{-3mm}
\label{fig:method}
\end{figure*}

%% file: sec/2_related_work.tex
\section{Related Work}
\label{sec:related_work}
% Embodied navigation~\cite{anderson2018evaluation} encompasses a variety of task categories. Among these, we focus on ObjectNav~\cite{savva2019habitat} and EQA~\cite{das2018embodied} in this paper.
% , where the integration of visual and linguistic information is essential for navigating and interacting effectively in complex environments.

\subsection{Object Goal Navigation}
% 介绍on的方法，end2end，模块化
The ObjectNav~\cite{savva2019habitat} research landscape can be categorized into closed-world and open-world settings. Closed-world ObjectNav~\cite{mousavian2019visual, shen2019situational, campari2020exploiting, xie2023implicit, zhang2023layout, du2023object, gupta2017cognitive, chaplot2020object, rudra2023contextual, campari2022online, chang2020semantic, zhu2022navigating} focuses on locating objects from a fixed set of object categories (e.g., ``chair''). Approaches include end-to-end methods~\cite{mousavian2019visual, shen2019situational, campari2020exploiting, xie2023implicit, zhang2023layout, du2023object} that map sensory inputs directly to actions, and modular architectures~\cite{gupta2017cognitive, chaplot2020object, rudra2023contextual, campari2022online, chang2020semantic, zhu2022navigating} that separate perception, mapping, and control. While effective in controlled settings, these methods are limited by their reliance on predefined object classes. To address this limitation, Open Vocabulary Object Goal Navigation (OVON)~\cite{yokoyama2024hm3d} allows users to specify any object using natural language (e.g., ``find a television''), enabling greater adaptability to real-world scenarios. Although current ObjectNav models are effective at finding target objects, they often struggle to explain their navigation decisions.

% \vspace{-1mm}
\subsection{Embodied Question Answering}
% \vspace{-0mm}
% 先介绍eqa的定义，然后介绍一下著名的eqa数据集，最后再说我们设计的eqa数据集和它们相比优势是什么
EQA~\cite{das2018embodied,huang2024drivemm} tasks require agents to explore 3D environments and answer visually grounded questions, integrating visual perception, exploration, and language understanding. For example, agents need to navigate physical spaces to answer questions such as ``What color is the car?''
The EQA task developed by the MT-EQA dataset~\cite{yu2019multi} introduces scenarios involving multiple objects and locations, thereby increasing its complexity.
Furthermore, the MP3D-EQA dataset~\cite{wijmans2019embodied} advances the field by incorporating photorealistic environments with point cloud perception, challenging agents to interact with detailed 3D spaces. Recently, the HM-EQA dataset~\cite{ren2024explore} refines exploration strategies with an ``explore until confident'' approach, promoting efficient data gathering for accurate decision-making. These aforementioned developments collectively enhance the capabilities of agents in EQA tasks.
Based on EQA datasets, some studies~\cite{zheng2024towards, zhang2024uni} advance the field by unifying navigation and EQA tasks into a comprehensive model.
% This integration enhances agents' comprehension and interaction with their environment, leading to more robust and versatile navigation solutions, while also fostering the development of more generalized and adaptable models.

%% file: sec/3_method.tex
\section{Methodology}
\label{sec:method}
% We present \textbf{RoboTron-Nav}, a novel and unified framework for embodied navigation that seamlessly integrates the essential components of \textbf{P}erception, \textbf{P}lanning, and \textbf{P}rediction. First, we provide an overview of RoboTron-Nav's architecture, detailing its various modules and highlighting its central component: the \textbf{Adaptive 3D-aware History Sampling} strategy, which is crucial for effectively and efficiently utilizing historical visual information in navigation. Next, we introduce an innovative method for extending navigation datasets with EQA pairs that explicitly model the decision-making processes involved in navigation. Finally, we present a \textbf{Multitask Collaboration} strategy that bolsters navigation capabilities by enabling joint training on both navigation and EQA tasks, allowing agents to leverage perception and planning to improve their prediction capabilities.

In this section, we present {RoboTron-Nav}, a unified framework for embodied navigation that integrates {p}erception, {p}lanning, and {p}rediction. We first overview RoboTron-Nav and its core {adaptive 3D-aware history sampling} strategy (Sec.~\ref{sec:model}). Next, we introduce a method to augment navigation datasets with EQA pairs for explicit decision modeling (Sec.~\ref{sec:dataset}). Finally, we describe a {multitask collaboration} strategy (Sec.~\ref{sec:mc}) that enhances navigation by jointly training on navigation and EQA tasks.
% \vspace{-1mm}

\subsection{Model}
\label{sec:model}
As shown in Fig.~\ref{fig:method}, RoboTron-Nav integrates three components: a visual encoder for current observations, adaptive 3D-aware history sampling for context modeling, and an LLM for action and answer prediction.
% The visual encoder processes the current frame through a two-component architecture: the Vision Transformer (ViT)~\cite{radford2021learning} extracts 2D image features in parallel with UVFormer~\cite{liu2024robouniview} encoding 3D spatial structures, jointly emphasizing real-time perceptual cues.
% After that, the \textbf{Adaptive 3D-aware History Sampling} strategy processes historical frames by adaptively sampling key segments that preserve critical motion dynamics while reducing resource consumption. 
% The processed visual tokens are concatenated with linguistic instruction tokens derived from text prompts, and this unified sequence of tokens is then input into the LLM. Consequently, the LLM decodes them to generate executable navigation actions and natural language answers. 
% As a result, RoboTron-Nav establishes explicit connections between perceptual cues and decision rationale.

% The hidden outputs are then projected to action and value predictions.

\subsubsection{Visual Encoder}
% ok:加一句当前帧的重要性
Our visual encoder integrates 2D and 3D features to represent the current observation frame, with a dedicated emphasis on real-time perceptual cues. It consists of two specialized components: a ViT~\cite{radford2021learning} for 2D feature extraction, and a UVFormer~\cite{liu2024robouniview} for 3D spatial encoding.

First, we utilize the CLIP-pretrained ViT~\cite{radford2021learning} as the image backbone for 2D feature extraction:
\begin{align}
X_{t} & = \operatorname{ViT}\left(I_{t}\right),
\end{align}
where $I_{t} \in \mathbb{R}^{ H \times W \times 3}$ denotes the RGB image, and $X_{t} \in \mathbb{R}^{n_{\text{img}} \times c}$ represents their corresponding 2D features at timestep $t$. Here, $n_{\text{img}}$ indicates the number of image patches while $c$ denotes the feature dimension.

% todo：这里应该补一句使用头部视角和腕部视角，在补充材料里补充
Additionally, inspired by~\cite{liu2024robouniview}, we implement multi-perspective alignment through unified view projection:
% (e.g., head and wrist perspectives) through unified view projection:
\begin{align}
UV_{t} & = \operatorname{UVFormer}\left(Q, X_{t}, Cam\right).
\label{eq:uvformer}
\end{align}
Here, the UVFormer~\cite{liu2024robouniview} module processes three inputs: image features $X_{t}$, camera parameters $Cam$, and learnable queries $Q$ that encode spatial positions and semantic features within the robot's 3D workspace.
 % (See \cite{liu2024robouniview} for architectural details)
The output $UV_{t} \in \mathbb{R}^{n_{\text{UV}} \times c}$ aggregates multi-view visual information through a structured 3D feature volume, where $n_{\text{UV}}$ denotes the number of 3D visual tokens.
See the supplementary material for more details.

\subsubsection{Adaptive 3D-aware History Sampling}
\label{sec:adaptive}
Utilizing all observed RGB frames can substantially mitigate the issue of catastrophic knowledge forgetting, which is essential for long-term navigation. However, due to GPU memory constraints and high computational costs, storing all image features extracted from these frames is often impractical. Additionally, we have observed a significant amount of redundancy in observations during navigation, as the agent frequently revisits the same locations multiple times.
% ok: 滑动窗口放后面并减弱，将位置编码加上去
% To address this issue, we propose a method for selecting those RGB frames that do not overlap in spatial positions to serve as historical RGB frames, thereby reducing observational redundancy. 
% Furthermore, we use the 3D positions of agents as the positional encoding for image features to enhance 3D perception. 
% This approach enhances historical semantic features with 3D trajectory information, preventing redundant exploration of the same position.
% This approach converts the dense sequence of frames into sparse sequence of features, reducing resource consumption while preserving visual information at key locations.
% We have observed a significant amount of redundancy in observations during navigation, as the agent frequently revisits the same locations multiple times. Utilizing all observed RGB frames can substantially mitigate the issue of catastrophic knowledge forgetting, which is essential for long-term navigation tasks. However, due to GPU memory constraints and high computational costs, storing all image features extracted from these frames is often impractical.
{To address this issue, we propose an adaptive 3D-aware history sampling strategy, which selects sparse and organized features based on historical position and viewpoint information, thereby reducing observational redundancy.}
% \textcolor{red}{To address this issue, we propose selecting historical RGB frames as valid observations when they are either non-adjacent in spatial position or, if spatially adjacent, captured from different viewpoints. Specifically, we determine whether frames are adjacent in spatial position based on their pairwise relative distances, and whether they are captured from different viewpoints based on their semantic similarities. By independently controlling the density and diversity of historical sampling in both the spatial and semantic domains, our approach effectively reduces observational redundancy while preserving critical information.}
% 
As shown in Algorithm~\ref{alg:adaptive_sampling_revised}, \textcolor{black}{we sample historical observations by using relative distance to determine spatial adjacency and semantic similarity to identify similar viewpoints.}
Our {adaptive 3D-aware history sampling} strategy operates through three sequential phases:  {initialization},  {3D-aware sampling}, and  {adaptive padding}. 

% Furthermore, we use the 3D positions of agents as the positional encoding for image features to enhance 3D perception. 
% This approach enhances historical semantic features with 3D trajectory information, preventing redundant exploration of the same position.
% To address this issue, we propose 3D-aware history features that use the 3D positions of the agent to enhance historical semantic features with 3D trajectory information, preventing redundant exploration of the same position. Furthermore, we propose a 3D-aware history sampling method for selecting those frames that do not overlap in spatial positions to serve as historical frames, thereby reducing observational redundancy. 

\begin{algorithm}
\caption{Adaptive 3D-aware History Sampling}
\label{alg:adaptive_sampling_revised}
\begin{algorithmic}[1]
\REQUIRE Current step $t$, Window size $W$, Relative Position Threshold $\epsilon$, Semantic Similarity Threshold $\tau$
\ENSURE Historical features $\mathbf{V}$, Relative positions $\mathbf{P}$

\STATE $\mathbf{V} \gets \emptyset$, $\mathbf{P} \gets \emptyset$
\STATE $\mathbf{G} \gets \text{GetAllObs}()$

\IF{$\mathbf{G} = \emptyset$}
    \STATE \# Initialize with dummy data
    \STATE \textbf{return} $\mathbf{V}$, $\mathbf{P}$
\ENDIF
\STATE \# Current frame as reference
\STATE ${p}^{\text{ref}} \gets \mathbf{G}[-1].{p}$
\STATE $k \gets 0$
\STATE \# From current time $t$ to 0
\FOR{$i \gets |\mathbf{G}|-1$ \textbf{to} $0$}
    \STATE ${p}_i^{\text{rel}} \gets \mathbf{G}[i].{p} - {p}^{\text{ref}}$
    
    \IF{$i < |\mathbf{G}|-1$ \AND $\exists\, j \in [0, k]:\ \|{p}_i^{\text{rel}} - \mathbf{P}[j]\|_2 < \epsilon \;$ \AND $\; \cos\left(\mathbf{G}[i].{v},\, \mathbf{V}[j]\right) > \tau$}
        \STATE \# Skip redundant frames
        \STATE \textbf{continue}
    \ENDIF
    % ok：这里不需要vit
    \STATE $\mathbf{V} \gets \mathbf{V} \oplus \mathbf{MaxPool}(\mathbf{G}[i].{v})$
    \STATE $\mathbf{P} \gets \mathbf{P} \oplus \mathbf{p}_i^{\text{rel}}$
    \STATE $k \gets k + 1$
    
    \IF{$k = W$} 
        \STATE \textbf{break} 
    \ENDIF
\ENDFOR

\WHILE{$k < W$} 
    \STATE \# Padding with last valid frame
    \STATE $\mathbf{V} \gets \mathbf{V} \oplus \mathbf{V}[-1]$
    \STATE $\mathbf{P} \gets \mathbf{P} \oplus \mathbf{P}[-1]$
    \STATE $k \gets k + 1$
\ENDWHILE 

\end{algorithmic}
\end{algorithm}

\begin{figure*}[!htbp]
\centering
\includegraphics[width=\textwidth]{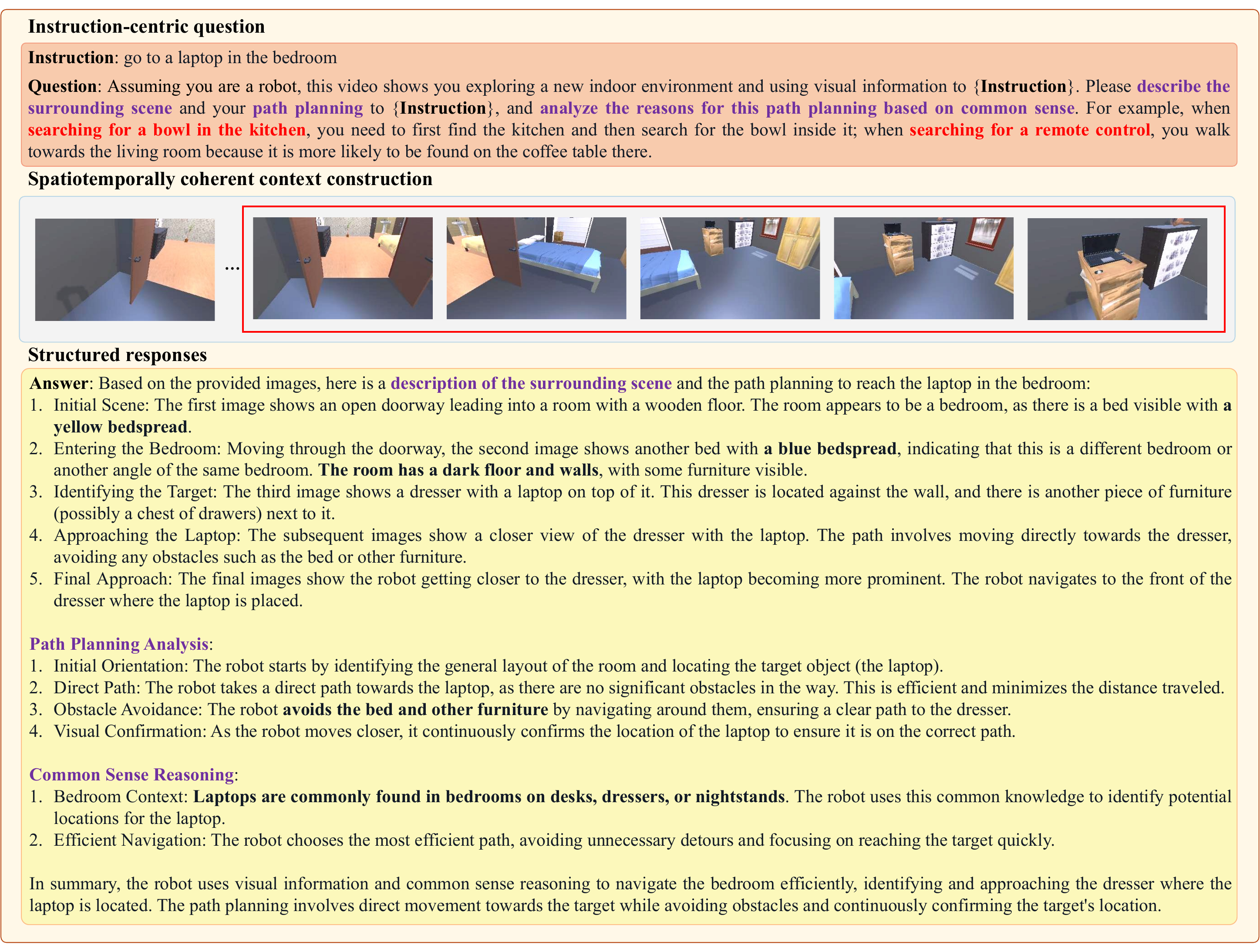}
\vspace{-2mm}
\caption{{EQA dataset extension framework.} The framework comprises three key stages: (1) standardized reasoning steps through instruction-centric question formulation, (2) decision context modeling via spatiotemporally coherent context construction, and (3) structured response generation integrating scene analysis, path planning, and commonsense reasoning. This hierarchical approach enables robots to systematically identify targets and formulate explainable navigation strategies while maintaining human-like decision transparency.}
\vspace{-3mm}
\label{fig:eqa}
\end{figure*}

% \paragraph{3D-aware history sampling.}
% Specifically, the {adaptive 3D-aware history sampling} strategy samples RGB frames in reverse chronological order, using the spatial position of the current frame as a reference. \textcolor{red}{We determine spatial adjacency using pairwise relative distances and distinguish different viewpoints using semantic similarities. RGB frames are retained if their pairwise relative distances exceed a predefined threshold; otherwise, only those with semantic similarities below a predefined threshold are kept.}

\textbf{Initialization.} The {initialization} phase (Lines~1-6) sets up the operational parameters by accepting the input parameters (current step $t$, window size $W$, relative position threshold $\epsilon$, and semantic similarity threshold $\tau$) and initializing empty buffers for historical features $\mathbf{V}$ and relative positions $\mathbf{P}$, respectively. 
% It handles edge cases through dummy data initialization when no historical frames exist, ensuring algorithmic robustness. 
\textcolor{black}{The \textit{GetAllObs} function returns all historical observations (i.e., $X_{0}, \ldots, X_{t-1}$ in Fig.~\ref{fig:method}), which are stored in $\mathbf{G}$ as a First-in-First-out (FIFO) queue, where $\mathbf{G}[0]$ corresponds to the initial time and $\mathbf{G}[-1]$ to the current time.} At time step $i$, $\mathbf{G}[i]$ includes historical features $\mathbf{G}[i].{v}$ paired with their corresponding agent absolute positions $\mathbf{G}[i].{p} = (x, y, z)$.
% ok：这里缺一句对GetAllPastObs函数的介绍
% ok：补充G[0]是初始时刻，G[-1]是当前时刻

\textbf{3D-Aware Sampling.} The core {3D-aware sampling} phase (Lines~8-23) processes all historical observations in $\mathbf{G}$ in reverse chronological order, from newest to oldest. 
First, it uses the robot's absolute position ${p}^{\text{ref}}$ at the current step $t$ as the spatial reference to compute relative coordinates ${p}_i^{\text{rel}}$ of historical features $\mathbf{G}[i].{v}$. 
\textcolor{black}{
Then, using the relative position threshold $\epsilon$ and the semantic similarity threshold $\tau$, the method dynamically filters out redundant visual observation. Specifically, for each candidate feature at time step $i$, we compare it with all previously selected historical features at time step $j$: if $|{p}_i^{\text{rel}} - \mathbf{P}[j]| < \epsilon$ (i.e., adjacent in spatial position) and $\cos\left( \mathbf{G}[i].{v}, \mathbf{V}[j] \right) > \tau$ (i.e., captured from similar viewpoints), feature $i$ is excluded from the historical set. By adjusting the value of $\epsilon$ and $\tau$, we can control the density and diversity of historical sampling in the spatial and semantic domains, respectively. 
% Smaller values of $\epsilon$ result in denser spatial sampling, allowing more continuous historical information to be retained, while smaller values of $\tau$ lead to stricter semantic filtering, ensuring greater diversity among the sampled features.
}
% ok: 加一句补充说明小于阈值删除，大于阈值保留
% ok:说明阈值和步长的关系，可以做到均匀采样
% todo：在补充材料里介绍maxpool函数
Concurrently, historical features $\mathbf{G}[i].{v}$ undergo dimensionality reduction via a $\mathbf{MaxPool}$ operator before being stored to $\mathbf{V}$. This phase terminates when the collected samples reach the target window size $W$ or $\mathbf{G}$ is empty.

\textbf{Adaptive Padding.} The {adaptive padding} phase (Lines~24-28) ensures fixed-length outputs by replicating the last valid entries whenever the historical features buffer $\mathbf{V}$ does not reach the designated window size $W$. Through cyclic duplication of $\mathbf{V}[-1]$ and $\mathbf{P}[-1]$, it maintains temporal coherence while fulfilling model input requirements. 
% Our \textbf{Adaptive 3D-aware History Sampling} algorithm efficiently removes spatial redundancies while preserving crucial temporal patterns necessary for navigation reasoning.

% ok:需要说明，robot的高度不会变化，因此只有x和y。此外，只有历史帧使用pe，当前时刻不使用
% \vspace{4mm}
Afterwards, we introduce {position-enhanced historical features} that utilize the agent's positions to augment historical semantic features with trajectory information, preventing redundant exploration of the same location.
%\noindent \textbf{Position-enhanced historical features.}
% \noindent{\textbf{Position-enhanced historical features.}} 
% Position encoding is essential for transformer-based architectures, providing the contextual information needed to understand the order and relative positions of input data. 
For obtained historical features $\mathbf{V}$, positional information is critical for recording the agent's historical trajectory and planning efficient paths for future exploration.
% For historical features $\mathbf{V}$, position embedding is crucial for maintaining the temporal sequence of observations, allowing the model to discern patterns and dependencies over time.
Hence, we obtain position-enhanced historical features by using the agents' relative positions $\mathbf{P}$ as positional encodings for the historical features $\mathbf{V}$. This integration enriches spatial semantics with 3D trajectory information, ensuring observation consistency and effectively preventing redundant exploration of previously visited locations.
As agents move on the ground, their positions ${p}$ involve constantly changing $x$ and $y$ coordinates, while the $z$ coordinate, representing height, remains constant. Consequently, we focus on encoding the agent's positions along the $x$ and $y$ axes to effectively capture their 3D trajectory.
% Consequently, this dynamic demands distinct but interrelated representations to accurately capture the complexity of the data.
Inspired by~\cite{vaswani2017attention}, we propose a 2D encoding through axis-separated frequency projections:  
\begin{equation}
\omega_k = e^{-2k(\log(10000))/d}, \quad d = c/2,
\end{equation}
\begin{equation}
PE_x = \Big[\sin(x\omega_{\lfloor m/2 \rfloor}), \cos(x\omega_{\lfloor m/2 \rfloor}) \Big]_{m=0}^{d-1},
\end{equation}
\begin{equation}
PE_y = \Big[\sin(y\omega_{\lfloor n/2 \rfloor}), \cos(y\omega_{\lfloor n/2 \rfloor}) \Big]_{n=0}^{d-1},
\end{equation}
\begin{equation}
PE(x,y) = PE_x \oplus PE_y,
\end{equation}
% \begin{align}
% \omega_k = e^{-2k(\log(10000))/d}, \quad d = c/2, \\
% PE_x = \Big[\sin(x\omega_{\lfloor m/2 \rfloor}), \cos(x\omega_{\lfloor m/2 \rfloor}) \Big]_{m=0}^{d-1}, \\
% PE_y = \Big[\sin(y\omega_{\lfloor n/2 \rfloor}), \cos(y\omega_{\lfloor n/2 \rfloor}) \Big]_{n=0}^{d-1}, \\
% PE(x,y) = PE_x \oplus PE_y,
% \end{align}
where $\oplus$ operation merges $x$/$y$-axis encodings into a unified spatial. 
% while preserving computational efficiency through parallel axis processing.
% ok：对最终得到的V和P，经过fc之后相加得到V
We add the positional encoding $PE(x,y)$ to each historical feature in $\mathbf{V}$, and then pass these features through a fully connected layer to obtain the position-enhanced historical features ${V_{p}} \in \mathbb{R}^{n_{\text{his}} \times c}$
% For each historical feature in $\mathbf{V}$, we calculate its position encoding $PE(x,y)$. By adding these features together and passing the result through a fully connected layer, we obtain the {position-enhanced historical features} ${V_{p}} \in \mathbb{R}^{n_{\text{his}} \times c}$
% by enriching the historical features with positional information
, where $n_{\text{his}}$ denotes the number of historical features.

\subsubsection{LLM}
% ok：这里要写language tokens
Finally, we integrate the {position-enhanced historical features} ${V_{p}}$ with the current observations ${UV_{t}}$ and ${X_{t}}$ from the visual encoder to create visual tokens for the LLM. In this framework, ${UV_{t}}$ and ${X_{t}}$ capture the agent's ongoing environment, while ${V_{p}}$ offers context from past observations, effectively filtering out redundant positions. Similarly, input instructions or questions are transformed into language tokens ${L} \in \mathbb{R}^{n_{\text{L}} \times c}$ via a tokenizer, where $n_{\text{L}}$ denotes the length of the language tokens. These visual and language tokens are combined and fed into the LLM.
% todo：如何在测试的时候生成answer我还没想出来
% todo：在补充里介绍两个head的结构
The LLM utilizes its abilities in multimodal alignment and comprehension to process these tokens, decoding the outputs through specialized heads. For navigation datasets, it employs an action head to generate the executable action, while for EQA datasets, it uses an answer head (i.g., LLM head) to produce the natural language answer to input questions.
% This integrated approach effectively combines visual and linguistic data, enhancing the model's ability to understand and execute a wide range of complex tasks.
% Please refer to the supplementary material for more details.

\subsection{Embodied Question Answering}
\label{sec:dataset}
Current navigation models merely mimic human navigation trajectories, lacking deep thinking and task planning similar to human cognitive processes. 
For instance, humans typically first identify related rooms, then locate furniture, and finally search for specific objects. 
% The lack of this ability leads to reduced navigation efficiency and poor transferability performance.
This lack of interpretability stems from traditional navigation datasets, which mainly focus on recording coordinate trajectories while neglecting the hierarchical reasoning used by humans. 
To address this issue, we have developed a new method for extending navigation datasets with EQA pairs that explicitly models the decision-making process involved in navigation.

As shown in Fig.~\ref{fig:eqa}, we instruct GPT-4o~\cite{hurst2024gpt} to generate perception of the surrounding environment, navigation planning, and navigation analysis for each navigation objective.
It consists of three stages: instruction-centric question, spatiotemporally coherent context construction, and structured responses.

% Current navigation models are proficient at locating target objects, but they often struggle to explain their reasoning behind path choices. This lack of interpretability stems from traditional navigation datasets, which mainly focus on recording coordinate trajectories while neglecting the hierarchical reasoning used by humans. For instance, humans typically first identify rooms, then locate furniture, and finally search for specific objects. To address this issue, we have developed a new method for constructing Embodied Question Answering (EQA) datasets that explicitly models the decision-making process involved in navigation. As shown in fig~\ref{fig:eqa}, this approach has three stages: (1) instruction-centric question that standardize reasoning steps, (2) spatiotemporally coherent context construction, (3) structured responses generation incorporating scene analysis, path planning, and commonsense justifications.

% 这个例子出自episode_id: 033173, traj_id: 0，可在spoc_gpt4o_analysis.ipynb的vis中找到
\noindent{\textbf{Instruction-centric question.}} We construct an instruction-centric question to elicit comprehensive first-person navigation analysis (e.g., ``\textit{go to a laptop in the bedroom}'') using two synergistic mechanisms: role specification (``\textit{assuming you are a robot}'') and exemplar-based prompting (two navigation reasoning and planning examples, see Fig.~\ref{fig:eqa}). This ensures structured outputs covering: (1) scene description (e.g., bedroom furniture layout), (2) path planning (e.g., obstacle avoidance), and (3) common sense reasoning (e.g., typical laptop locations).

% \paragraph{Instruction-centric question.} We construct an instruction-centric question that helps achieve navigation objectives (e.g., ``\textit{go to a laptop in the bedroom}'') through two synergistic mechanisms: role specification (``\textit{assuming you are a robot}'') and exemplar-based prompting (``\textit{searching for a bowl in the kitchen}''). This dual approach ensures structured outputs that contain three essential components: (1) surrounding scene description (e.g., recognizing bedroom furniture layouts), (2) path planning analysis (e.g., creating obstacle-avoidance trajectories), and (3) common sense reasoning (e.g., prior knowledge about where laptops are typically placed).

\noindent{\textbf{Spatiotemporally coherent context construction.}} To ensure alignment with the {adaptive 3D-aware history sampling} strategy established in Sec.~\ref{sec:adaptive}, we systematically select the final $W$ frames from each navigation trajectory. This frame range captures critical target-approaching phases while maintaining temporal consistency with upstream visual processing modules. By adopting this temporal window, we construct spatiotemporally coherent contexts that encode successful navigation patterns and preserve motion dynamics essential for trajectory analysis.

\noindent{\textbf{Structured responses.}} Building on the prepared inputs, we synthesize instruction-aligned multimodal prompts by combining textual questions with curated visual sequences. GPT-4o~\cite{hurst2024gpt} then generates structured responses in three phases. First, the surrounding scene description identifies architectural features (e.g., ``\textit{the room has a dark floor and walls}'') and object semantics (e.g., ``\textit{a yellow/blue bedspread}''). Second, path planning analysis devises hierarchical navigation strategies (e.g., ``\textit{enter bedroom → circumvent bed → final approach}''). Third, common sense reasoning incorporates human-environment interaction norms (e.g., ``\textit{laptops are commonly found on desks, dressers, or nightstands}''). Each phase operates sequentially, ensuring explicit alignment between perception, action planning, and contextual knowledge.

% todo：补充关于eqa数据集的详细信息
\subsection{Multitask Collaboration}
\label{sec:mc}
To address the limitations inherent in single-task learning, which only encompasses navigation tasks, we introduce a {multitask collaboration} strategy that significantly enhances navigation capabilities through joint training on both navigation and EQA tasks. As shown in Fig.~\ref{fig:method}, during the training phase, this strategy enables the model to learn from both tasks concurrently, with inputs consisting of navigational instructions and EQA questions. Consequently, the model can effectively utilize the perceptual and planning skills derived from EQA pairs to refine and improve its navigation prediction capabilities. By integrating these tasks, the model gains a comprehensive understanding of spatial relationships and decision-making processes, which are critical for accurate navigation.

During inference, the model focuses solely on predicting navigation actions, effectively leveraging the perceptual and planning abilities acquired during training for precise and efficient navigation. Please refer to the supplementary material for details on the training objective.

%% file: sec/4_experiments.tex
\begin{figure*}[!t]
\centering
\includegraphics[width=\textwidth]{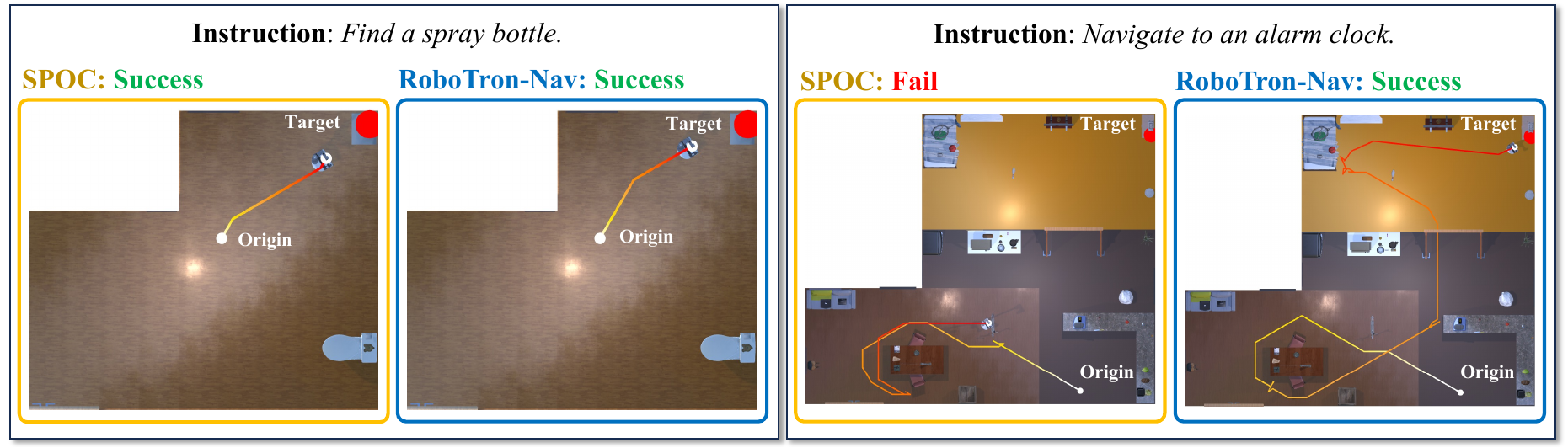}
\vspace{-7mm}
\caption{\textit{Left:} Both RoboTron-Nav and SPOC~\cite{ehsani2024spoc} generate the shortest path when the agent is near the target (e.g., in the same room). \textit{Right:} In contrast, SPOC~\cite{ehsani2024spoc} \textcolor{red}{fails} due to repeatedly searching along the same paths when distant from the target (e.g., in different rooms), whereas RoboTron-Nav \textcolor{green}{succeeds} by effectively avoiding revisiting areas and exploring new ones.}
\vspace{-2mm}
\label{fig:vis}
\end{figure*}

\section{Experiments}
\label{sec:experiments}
In this section, we first outline our experimental settings, then compare our proposed method with state-of-the-art techniques, and finally provide a detailed analysis of our approach, including qualitative results and ablation studies.

\subsection{Experimental Settings}
\label{sec:settings}
\paragraph{Dataset and Metrics.} We evaluate our proposed method on the $\mathrm{CHORES}$-$\mathbb{S}$ ObjectNav benchmark~\cite{ehsani2024spoc}. 
% ok：这里加一句eqa数据集的数据。这些都丢附录里，
% ok：说一下消融实验用的objnavroom
Besides, we extend the navigation dataset with EQA pairs for joint training, as described in Sec.~\ref{sec:dataset}.
We follow the official evaluation split of the $\mathrm{CHORES}$-$\mathbb{S}$ ObjectNav benchmark~\cite{ehsani2024spoc}, which contains 200 trajectories in 200 testing houses.
To comprehensively demonstrate the effectiveness of our method, we evaluate performance using three metrics: success rate (SR), episode-length weighted success (SEL)~\cite{eftekhar2023selective}, and percentage of rooms visited (\%Rooms)~\cite{ehsani2024spoc}. 
The additional explanations of the metrics and implementation details are provided in the supplementary materials.

\subsection{Performance Comparisons}
\label{sec:sota}
As shown in Table~\ref{tab:1}, we compare our method with PIRLNav~\cite{ramrakhya2023pirlnav}, JSRL~\cite{uchendu2023jump}, EmbSigLIP~\cite{ehsani2024spoc}, SPOC~\cite{ehsani2024spoc}, $\text{SPOC}^{*}$~\cite{ehsani2024spoc}, and RING~\cite{eftekhar2024one}. 
% Results are taken directly from the original papers or their supplementary materials; ``-'' indicates unavailable results.
\textcolor{black}{PIRLNav~\cite{ramrakhya2023pirlnav} and JSRL~\cite{uchendu2023jump} are first pretrained with imitation learning, and then fine-tuned via reinforcement learning with sparse rewards.} EmbSigLIP~\cite{ehsani2024spoc} denotes an EmbCLIP~\cite{khandelwal2022simple} model upgraded with a SigLIP~\cite{zhai2023sigmoid} backbone. SPOC~\cite{ehsani2024spoc} imitates shortest paths by modeling optimal routes, while $\text{SPOC}^{*}$~\cite{ehsani2024spoc} is trained on a larger set of expert trajectories. \textcolor{black}{RING~\cite{eftekhar2024one} introduces a universal indoor navigation policy transferable across diverse robots without additional fine-tuning.}
% RoboTron-Nav integrates perception, planning, and prediction by jointly training on navigation and EQA tasks via a multitask collaboration strategy, and processes historical frames using adaptive 3D-aware history sampling.

\begin{table}[!t]
    \centering
    \begin{tabular}{l|ccc}
    \hline
    \textbf{Method} & \textbf{SR $\uparrow$} & \textbf{SEL $\uparrow$} & \textbf{\%Rooms} \\ \hline
    PIRLNav~\cite{ramrakhya2023pirlnav} & 20.0 & {7.0} & - \\ 
    JSRL~\cite{uchendu2023jump} & 21.0 & {15.6} & - \\ 
    EmbSigLIP~\cite{khandelwal2022simple} & 36.5 & 24.5 & 42.2 \\ 
    SPOC~\cite{ehsani2024spoc} & 57.0 & {46.2} & 51.5 \\ 
    $\text{SPOC}^{*}$~\cite{ehsani2024spoc} & 60.0 & 30.5& - \\ 
    RING~\cite{eftekhar2024one} & 72.1 & \textbf{53.0} & - \\ 
    {RoboTron-Nav} & \textbf{81.1} & 43.9 & {67.8} \\ \hline
    \end{tabular}
    \vspace{-2mm}
    \caption{Performance comparison of different navigation methods on the $\mathrm{CHORES}$-$\mathbb{S}$ ObjectNav benchmark~\cite{ehsani2024spoc}.}
    \vspace{-3mm}
    \label{tab:1}
\end{table}

As presented in Table~\ref{tab:1}, \textcolor{black}{PIRLNav~\cite{ramrakhya2023pirlnav} and JSRL~\cite{uchendu2023jump} achieve relatively low SRs (20.0 and 21.0), which can be attributed to the challenges of fine-tuning with sparse reward signals in reinforcement learning.}
EmbSigLIP~\cite{khandelwal2022simple} also demonstrates limited navigation performance (SR 36.5, SEL 24.5).
SPOC~\cite{ehsani2024spoc} shows improvement, achieving an SR of 57.0 and a SEL of 46.2 by focusing on shortest path learning.
$\text{SPOC}^{*}$~\cite{ehsani2024spoc} further increases SR to 60.0 via additional expert trajectory training, though at the cost of longer paths, reducing SEL to 30.5.
\textcolor{black}{RING~\cite{eftekhar2024one} achieves a much higher SR of 72.1 and the best SEL of 53.0, demonstrating strong generalization and efficiency.}
RoboTron-Nav leverages adaptive 3D-aware history sampling and multitask collaboration to achieve the highest SR of 81.1, effectively utilizing historical data for improved path decision-making and enhancing environmental understanding through EQA tasks. These results demonstrate that RoboTron-Nav excels in long-term navigation, surpassing other methods by integrating planning across both navigation and EQA tasks.

\begin{table}[!t]
    \centering
    \resizebox{\columnwidth}{!}{%
    \begin{tabular}{l|ccc|ccc}
    \hline
    \textbf{Exp} & HO &SP& PEHF & \textbf{SR $\uparrow$} & \textbf{SEL $\uparrow$} & \textbf{\%Rooms} \\ \hline
    1& {\color[gray]{0.8} \ding{55}} & {\color[gray]{0.8} \ding{55}} & {\color[gray]{0.8} \ding{55}} & 31.3 & \textbf{16.7} & 65.6 \\ 
    2& \Checkmark & \Checkmark & {\color[gray]{0.8} \ding{55}} & 39.2 & 14.4 & {77.7} \\ 
    3& \Checkmark & {\color[gray]{0.8} \ding{55}} & \Checkmark & 40.3 & 9.2 & {81.1} \\ 
    4& \Checkmark & \Checkmark & \Checkmark & \textbf{42.5} & 16.0 & {62.3} \\ \hline
    \end{tabular}
    }
    \vspace{-2mm}
    \caption{Ablation study on 3D-aware history sampling, where HO, SP, and PEHF stand for historical observations, sampling, and position-enhanced historical features, respectively.}
    % \vspace{-3mm}
    \label{tab:window}
\end{table}

\begin{table}[!t]
    \centering
    \resizebox{\linewidth}{!}{%
    \begin{tabular}{l|c|cc|ccc} % 使用半角的竖线
    \hline
    \textbf{Exp} & benchmark & Navigation & EQA & \textbf{SR $\uparrow$} & \textbf{SEL $\uparrow$} & \textbf{\%Rooms} \\ \hline
    1 & ObjectNavRoom & \Checkmark & {\color[gray]{0.8} \ding{55}} & {42.5} & {16.0} & {62.3} \\
    2 & ObjectNavRoom & \Checkmark & \Checkmark & \textbf{47.2} & \textbf{21.9} & 65.8 \\ \hline
    3 & ObjectNav & \Checkmark & {\color[gray]{0.8} \ding{55}} & 72.5 & 33.5 & 68.4 \\ 
    4 & ObjectNav & \Checkmark & \Checkmark & \textbf{81.1} & \textbf{43.9} & 67.8 \\ \hline
    \end{tabular}%
    }
    \vspace{-2mm}
    \caption{Ablation on the multitask collaboration strategy.}
    \vspace{-3mm}
    \label{tab:hf}
\end{table}

\subsection{Qualitative Results}
To further illustrate the effectiveness of our unified framework for embodied navigation, some visualization results compared with SPOC~\cite{ehsani2024spoc} are illustrated in Fig.~\ref{fig:vis}.
In the left case, both RoboTron-Nav and SPOC~\cite{ehsani2024spoc} are capable of generating the shortest path when the agent is positioned close to the target, such as being in the same room. Both methods demonstrate a strong understanding of the environment, allowing them to navigate efficiently without significant difficulties. 
However, the right case illustrates that when the agent is farther from the target, such as in different rooms, SPOC~\cite{ehsani2024spoc} faces challenges by repeatedly searching the same paths, leading to inefficiencies in navigating complex environments.
% Consequently, SPOC~\cite{ehsani2024spoc} increases the risk of looping or missing optimal routes, significantly reducing its success rate.
% In contrast, RoboTron-Nav effectively avoids revisiting explored areas, thereby enabling it to systematically explore new paths and enhance navigation efficiency. Consequently, RoboTron-Nav can dynamically adjust its strategy to changing environments, making it highly effective for distant navigation and complex path optimization.
% In contrast, RoboTron-Nav avoids revisiting explored areas, allowing exploration of new paths and enhancing navigation efficiency, thereby adapting to changing environments and excelling in distant navigation.
In contrast, RoboTron-Nav avoids revisiting explored areas, enabling new path exploration and enhanced navigation efficiency, thus excelling in long-term navigation.

\cj{
\begin{table}[!t]
    \centering
    \begin{tabular}{l|ccc|ccc}
    \hline
    \textbf{Exp} &$W$ &$\epsilon$ & $\tau$ & \textbf{SR $\uparrow$} & \textbf{SEL $\uparrow$} & \textbf{\%Rooms} \\ 
    \hline
    1 & 20 & 0.1 & 0 & 32.6 & 13.2 & 74.8 \\
    2 & 40 & 0.1 & 0 & 34.8 & 13.0 & 74.7 \\
    3 & 60 & 0.1 & 0 & 42.1 & 15.2 & 75.0 \\
    4 & 80 & 0.1 & 0 & 38.9 & \textbf{16.7} & 74.3 \\
    5 & 100 & 0.1 & 0 & 35.2 & 14.3 & 72.3 \\
    \hline 
    6 & 60 & 0.05 & 0 & 41.5 & 10.3 & 79.9 \\
    7 & 60 & 0.1 & 0 & 42.1 & 15.2 & 75.0 \\
    8 & 60 & 0.15 & 0 & 41.3 & 12.6 & 77.8 \\
    9 & 60 & 0.2 & 0 & 40.5 & 13.7 & 72.1 \\
    \hline
    10 & 60 & 0.1 & 0.9 & 42.3 & 15.6 & 64.2 \\
    11 & {60} & {0.1} & {0.95} & \textbf{42.5} & {16.0} & {62.3} \\
    12 & 60 & 0.1 & 0.99 & 41.5 & 14.6 & 73.2 \\
    \hline
    \end{tabular}
    \vspace{-2mm}
    \caption{Analysis of different hyper-parameters for the 3D-aware history sampling strategy.}
    \vspace{-3mm}
    \label{tab:hyp}
\end{table}
}

\subsection{Ablation Studies}
% ok：这里需要补充消融实验在1M任务上完成
To demonstrate the effectiveness of different modules within our unified framework, we conduct ablation studies on the ObjectNavRoom benchmark~\cite{ehsani2024spoc}, which has only 1/5 the sample size of ObjectNav. See the supplementary materials for more details about the ObjectNavRoom.

% \begin{table}[!htbp]
%     \centering
%     \caption{Ablation on Window Size $W$.}
%     \label{tab:window}
%     \begin{tabular}{l|ccc}
%     \hline
%     $W$ & \textbf{SR $\uparrow$} & \textbf{SEL $\uparrow$} & \textbf{\%Room} \\ \hline
%     0   & 31.3 & \textbf{16.7} & 65.6 \\ 
%     20  & 32.6 & 13.2 & 74.8 \\ 
%     40  & 34.8 & 13.0 & 74.7 \\ 
%     60  & \textbf{42.1} & 15.2 & \textbf{75.0} \\ 
%     80  & 38.9 & \textbf{16.7} & 74.3 \\ 
%     100 & 35.2 & 14.3 & 72.3 \\ \hline
%     \end{tabular}
% \end{table}

\noindent \textbf{Does the adaptive 3D-aware history sampling help?}
As shown in Table~\ref{tab:window}, we conduct an ablation study on three key components in adaptive 3D-aware history sampling: historical observations (HO), sampling (SP), and position-enhanced historical features (PEHF). Exp.~1, which uses no historical observations ($W=0$), relies solely on current visual input and yields the lowest SR, underscoring the value of historical information. Comparing Exp.~2 and 4, removing PEHF leads to a 2.9 decrease in SR, indicating that positional information helps avoid redundant exploration and improves efficiency. Similarly, between Exp.~3 and 4, omitting sampling in historical observations (\textcolor{black}{i.e., $\epsilon=0$, $\tau=1$}) causes a 2.2 drop in SR, showing that sampling reduces redundancy and enhances navigation accuracy.

% \noindent\textbf{Does the position-enhanced historical features help?} 
% To investigate the efficacy of the position-enhanced historical features (PEHF) in enhancing model performance, we conducted an ablation study. As shown in Table~\ref{tab:hf}, the model trained with PEHF features (i.e., $\mathrm{V_{p}}$) in Exp. 2 demonstrates an increase of 2.9 in {SR} and 0.8 in {SEL} compared to the model trained with normal historical features $\mathrm{V}$ in Exp. 1. This result clearly demonstrates that incorporating PEHF significantly enhances navigation efficiency and success rate, validating the positive impact of position-enhanced historical features on model performance.

% \begin{table}[!htbp]
%     \centering
%     \caption{Ablation on 3D-aware History Features (HF).}
%     \label{tab:hf}
%     \begin{tabular}{l|ccc}
%     \hline
%     \textbf{Model} &\textbf{SR $\uparrow$} & \textbf{SEL $\uparrow$} & \textbf{\%Room} \\ \hline
%     RoboTron-Nav w/o HF& 39.2 & 14.4 & \textbf{77.7} \\ 
%     RoboTron-Nav &\textbf{42.1} & \textbf{15.2} & 75.0 \\ \hline
%     \end{tabular}
% \end{table}

\noindent \textbf{Does the multitask collaboration strategy help?}
We conduct an ablation study on the multitask collaboration strategy to evaluate the benefits of joint training on both navigation and EQA tasks. As demonstrated in Table~\ref{tab:hf}, Exp. 2, which additionally utilizes the EQA dataset during the training phase, achieves an increase of 4.7 in SR and 5.9 in SEL compared to Exp. 1, which only employs the navigation dataset. This result indicates that integrating the EQA dataset allows agents to leverage perception and planning to improve their prediction capabilities, thereby enhancing navigation abilities. Moreover, we have identified a remarkable improvement: Exp. 4, trained on the ObjectNav benchmark, which contains five times more samples than ObjectNavRoom, exhibits a \textbf{8.6} increase in SR compared to Exp. 3. This suggests that increasing the volume of EQA data can significantly enhance navigation performance.

\noindent \textbf{Effect of hyper-parameters.} \textcolor{black}{We investigate the performances using different values of the window size $W$, the relative position threshold $\epsilon$, and the semantic similarity threshold $\tau$ for the adaptive 3D-aware history sampling strategy. 
Here, $W$ controls the length of historical context, while $\epsilon$ and $\tau$ jointly determine the density and diversity of historical sampling in the spatial and semantic domains, respectively.
As shown in Table~\ref{tab:hyp}, we conducted comprehensive grid search experiments to analyze specifically how the window size $W$ (20-100), relative position threshold $\epsilon$ (0.05-0.2), and semantic similarity threshold $\tau$ (\cj{0.9}-0.99) affect performance. 
Increasing $W$ initially improves performance by providing richer historical information, with the best results at $W=60$ (around 6 meters). However, when $W$ exceeds 60, the inclusion of excessive history starts to overwhelm the model, leading to reduced performance. For spatial sampling, increasing $\epsilon$ up to 0.1 improves performance by reducing spatial redundancy through lowering the sampling frequency, with optimal results at $\epsilon=0.1$. However, further increases make sampling overly sparse and reduce performance. For semantic sampling, increasing $\tau$ up to 0.95 improves performance by preserving useful contextual information, with optimal results at $\tau=0.95$. However, further increases introduce excessive redundancy and reduce performance.
Therefore, we adopt $W=60$, $\epsilon=0.1$, and $\tau=0.95$ for our adaptive 3D-aware history sampling strategy.}

%% file: sec/5_conclusion.tex
\section{Conclusion}
\label{sec:conclusion}
In this paper, we present the {RoboTron-Nav} framework, which significantly advances language-guided visual navigation by effectively integrating perception, planning, and prediction through multitask collaboration strategy. Additionally, RoboTron-Nav adopts an adaptive 3D-aware history sampling strategy to enhance navigation by efficiently utilizing historical data to minimize redundancy. The framework's impressive 81.1\% success rate in ObjectNav on the $\mathrm{CHORES}$-$\mathbb{S}$ benchmark underscores its effectiveness. Although RoboTron-Nav has achieved remarkable results in simulated environments, there is still a long way to go to achieve truly human-like embodied navigation. Future work will address these limitations by exploring the integration of larger, more diverse datasets to further enhance RoboTron-Nav's generalization capabilities across various environments.

%% file: sec/X_suppl.tex
\clearpage
\setcounter{page}{1}
\maketitlesupplementary

\section{Implementation Details}
In this section, we elaborate on the more detailed implementations for RoboTron-Nav in Sec.~\ref{sec:method}.

\subsection{Model} 
\subsubsection{Vison Encoder} 
Following the settings in previous works~\cite{liu2024robouniview, yan2024robomm}, we input RGB images from both the head perspective $I_{t}^{head}$ and the wrist perspective $I_{t}^{wrist}$ into ViT~\cite{radford2021learning} to obtain the 2D features $X_{t}^{head}$ and $X_{t}^{wrist}$. Both $X_{t}^{head}$ and $X_{t}^{wrist}$ are then fed into UVFormer~\cite{liu2024robouniview} to construct multi-view 3D features $UV_{t}$. 

\textcolor{black}{We adopt UVFormer~\cite{liu2024robouniview} as our 3D occupancy predictor. As shown in Eq.~\eqref{eq:uvformer}, UVFormer takes the image features $X_t$, camera parameters $Cam$, and a set of learnable UniView queries $Q$ as input, and outputs a unified view representation $UV_t$. The query set $Q = \{Pos, Emb\}$ comprises positional encodings $Pos \in \mathbb{R}^{L \times B \times 3P}$ and learnable embeddings $Emb \in \mathbb{R}^{L \times B \times C}$. Here, $L$ and $B$ (both set to $20$) specify the 3D grid’s spatial layout within the robot’s workspace, and $P$ is the number of uniformly sampled points along the vertical axis of each pillar cell. Each pillar cell covers $0.05^2$ square meters on the ground and spans $0.5$ meters in height. $Emb^{l, b} \in \mathbb{R}^C$ encodes features for each pillar cell. The camera parameters $Cam$ correspond to $N$ different viewpoints. The unified view representation $UV_t \in \mathbb{R}^{L \times B \times C}$ integrates information from the entire $L \times B \times P$ 3D grid and serves as the basis for downstream occupancy prediction.}

In the navigation task, we adhere to the settings in~\cite{liu2024robouniview, yan2024robomm}, utilizing only the wrist perspective $X_{t}^{wrist}$ as the 2D feature $X_{t}$ to broaden the exploration view. 
Conversely, in the EQA task, we use the head perspective $X_{t}^{head}$ for the 2D feature $X_{t}$ to maintain a first-person perspective, as the EQA pairs are generated from this perspective.

\subsubsection{LLM} 
We utilize MPT\footnote{\url{https://huggingface.co/mosaicml/mpt-1b-redpajama-200b-dolly}} as our LLM, freezing the self-attention layers during training while fine-tuning the cross-attention layers. For the action head, we employ a multi-layer perceptron to map the final hidden states produced by the LLM from the $c$-dimensional space to the action space of the $\mathrm{CHORES}$-$\mathbb{S}$ ObjectNav benchmark~\cite{ehsani2024spoc}.
For the answer head (i.e., LLM head), we apply the \textit{argmax} operation on the logits output by the LLM to decode the answer.

\subsection{Training Objective}
To achieve \textbf{Multitask Collaboration}, we design a unified loss function that jointly optimizes navigation actions, question answering, and 3D occupancy through modality-specific components:
\begin{equation}
\mathcal{L} = \mathcal{L}_{\text{action}} + \mathcal{L}_{\text{answer}} + \lambda_{\text{occ}} \mathcal{L}_{\text{occ}},
\end{equation}
where $\mathcal{L}_{\text{action}}$, $\mathcal{L}_{\text{answer}}$, and $\mathcal{L}_{\text{occ}}$ denote navigation action prediction loss, embodied question answering loss, and 3D occupancy prediction loss respectively. 
The term $\lambda_{\text{occ}}$ is the weight coefficient for the occupancy loss. 

\noindent\textbf{Action prediction Loss.} We utilize behavior cloning to train the navigation model. Given an expert trajectory  $\tau=\left(\hat{a_{0}}, \cdots, \hat{a_{T}}\right)$, we use the cross-entropy loss for action prediction. The loss for the trajectory is as follows:
\begin{equation}
\mathcal{L}_{\text{action}} = \sum_{t=1}^{T}\left(CE(a_t, \hat{a}_t)\right),
\end{equation}
where $a_t$ denotes the predicted action and $\hat{a}_t$ the ground-truth (GT) demonstration at timestep $t$.

\noindent\textbf{Question answering loss.} Given the GT answer $y_{1: K}$ of the input question {with the length of $K$}, we optimize the generated answer token probabilities by a conventional cross-entropy loss:
\begin{equation}
\label{eq:6}
L_{\text{answer}}=-\sum_{k=1}^{K} \log \left(p\left(y_{k} \mid y_{1: k-1}\right)\right).
\end{equation}

\noindent\textbf{Occupancy loss.} Following the approach used in previous works~\cite{liu2024robouniview, yan2024robomm}, we utilize a standard cross-entropy loss function, denoted as ${L}_{\text{occ}}$, on the generated 3D volume.

\section{Experimental Settings}
\subsection{Dataset and Metrics}
\subsubsection{Dataset} 
\textcolor{black}{We selected the $\mathrm{CHORES}$-$\mathbb{S}$ benchmark for its complex indoor environments (10K rooms) and diverse object categories (15 types), allowing for comprehensive navigation testing. }
The $\mathrm{CHORES}$-$\mathbb{S}$ ObjectNav benchmark~\cite{ehsani2024spoc} includes 15 object categories and annotates 99k trajectories within 10k training houses, among 5M expert trajectory frames in the AI2-THOR simulated environment~\cite{kolve2017ai2}. 
For the $\mathrm{CHORES}$-$\mathbb{S}$ ObjectNav benchmark~\cite{ehsani2024spoc}, we extend each trajectory with EQA pairs. As a result, we collect 99k EQA pairs as the corresponding EQA dataset for joint training.
The $\mathrm{CHORESNAV}$-$\mathbb{S}$ ObjectNavRoom benchmark~\cite{ehsani2024spoc} is similar to the ObjectNav benchmark but involves smaller trajectories. This benchmark includes 15 object categories and annotates 21k trajectories within 2k training houses out of 1M expert trajectory frames.
Additionally, the ObjectNavRoom benchmark uses more diverse instructions, describing both the object's category and its room type simultaneously, such as ``Find a vase in the living room.'' In contrast, the ObjectNav benchmark specifies only the object's category, such as ``Find a vase.''
Similarly, we extend each trajectory in the ObjectNavRoom benchmark with EQA pairs, collecting 21k EQA pairs as the corresponding EQA dataset for ablation studies.

The action space of the ObjectNav and ObjectNavRoom benchmarks~\cite{ehsani2024spoc} includes 20 actions: Move Base ($\pm 20 \mathrm{~cm}$); Rotate Base ($\pm 6^{\circ}$, $\pm 30^{\circ}$); Move Arm (x, z) ($\pm 2 \mathrm{~cm}$, $\pm 10 \mathrm{~cm}$); Rotate Grasper ($\pm 10^{\circ}$); pickup; dropoff; done with subtask; and terminate.

\subsubsection{Metrics} 
\textbf{Success rate (SR)} is defined as the proportion of episodes deemed successful, which occurs when the agent executes the ``end'' action and the distance to the target, any instance of the category, is within a specified threshold (e.g., $2m$).
\textbf{Episode-length weighted success (SEL)}~\cite{eftekhar2023selective} is a metric used to evaluate the efficiency of an agent’s navigation. It compares the shortest possible path to the agent’s actual path, calculated as:
\begin{equation}
\frac{1}{N} \sum_{i=1}^{N} S_{i} \frac{w_{i}}{\max \left(w_{i}, e_{i}\right)},
\end{equation}
where \( w_{i} \) represents the shortest possible episode length to the target object, \( e_{i} \) is the episode length produced by the agent, and \( S_{i} \) is a binary indicator that denotes success for episode \( i \).
\textbf{Percentage of rooms visited (\%Rooms)} is a metric that measures the proportion of distinct rooms an agent successfully visits during navigation relative to the total number of rooms available in the environment. This metric reflects the agent's exploratory capability and efficiency in covering different areas within a given space.

\subsection{Traing Strategy}
Here, we describe the model hyper-parameters and training details of RoboTron-Nav. 

\subsubsection{Model Hyper-parameters}
In the visual encoder, the number of image patches \( n_{\text{img}} \) is set to 64, the number of multi-view vision tokens \( n_{\text{uv}} \) is 400, and the feature dimension \( c \) is 1024. 
In the {adaptive 3D-aware history sampling} strategy, the window size \( W \) is 60, the proximity threshold \( \epsilon \) is 0.1, and the number of historical frames \( n_{\text{his}} \) is 60. For the $\mathbf{MaxPool}$ operator, we use an adaptive max pooling function to reduce the number of tokens in the historical features $\mathbf{G}[i].\mathbf{v}$ to 1.
In the LLM, the number of language tokens \( n_{\text{L}} \) corresponds to the length of input instructions and questions, respectively.

\subsubsection{Training Details}
We train the entire model with the AdamW optimizer using 8 A100 GPUs (80 GB memory per GPU), with a batch size of 48 per GPU, resulting in a total batch size of 384 for 5 epochs. A cosine learning rate strategy is employed, where the learning rate is initially set to $1 \times 10^{-4}$ and finally decays to $1 \times 10^{-6}$.
We evaluate checkpoints every 0.5 epoch starting from the 3rd epoch and report the metrics for the checkpoint with the highest SR on the evaluation split.

\subsubsection{Training Efficiency and Convergence Stability}
In terms of training efficiency, multitask training requires approximately twice as much time as single-task training. Additionally, both approaches demonstrate stable loss reduction and typically converge by the fifth epoch.

\section{Qualitative Results}
\begin{figure*}[!htbp]
\centering
\includegraphics[width=0.65\textwidth]{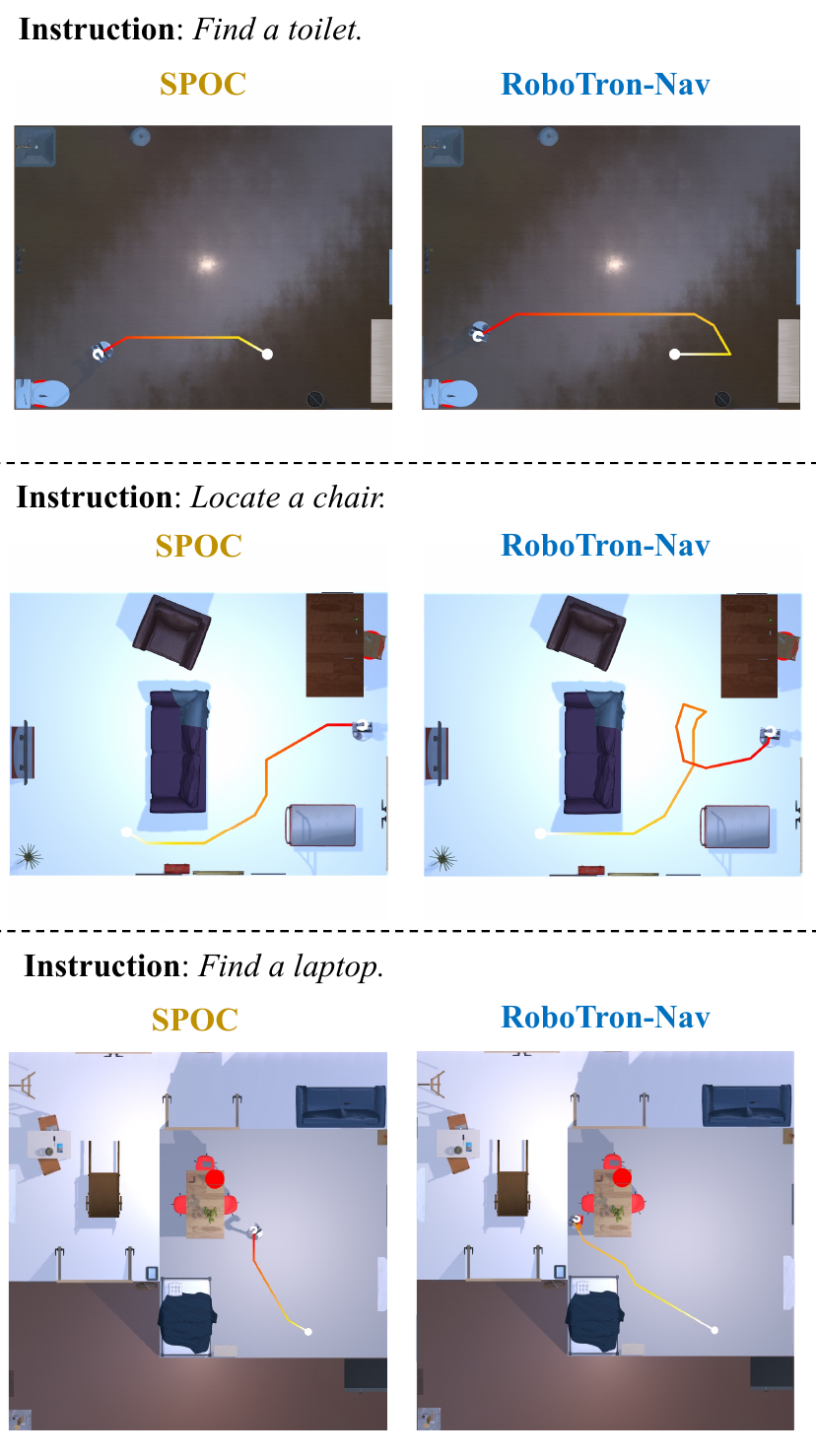}
\caption{Qualitative comparison of trajectories generated by SPOC~\cite{ehsani2024spoc} and RoboTron-Nav in the same room.}
\label{fig:short}
\end{figure*}

\begin{figure*}[!htbp]
\centering
\includegraphics[width=0.65\textwidth]{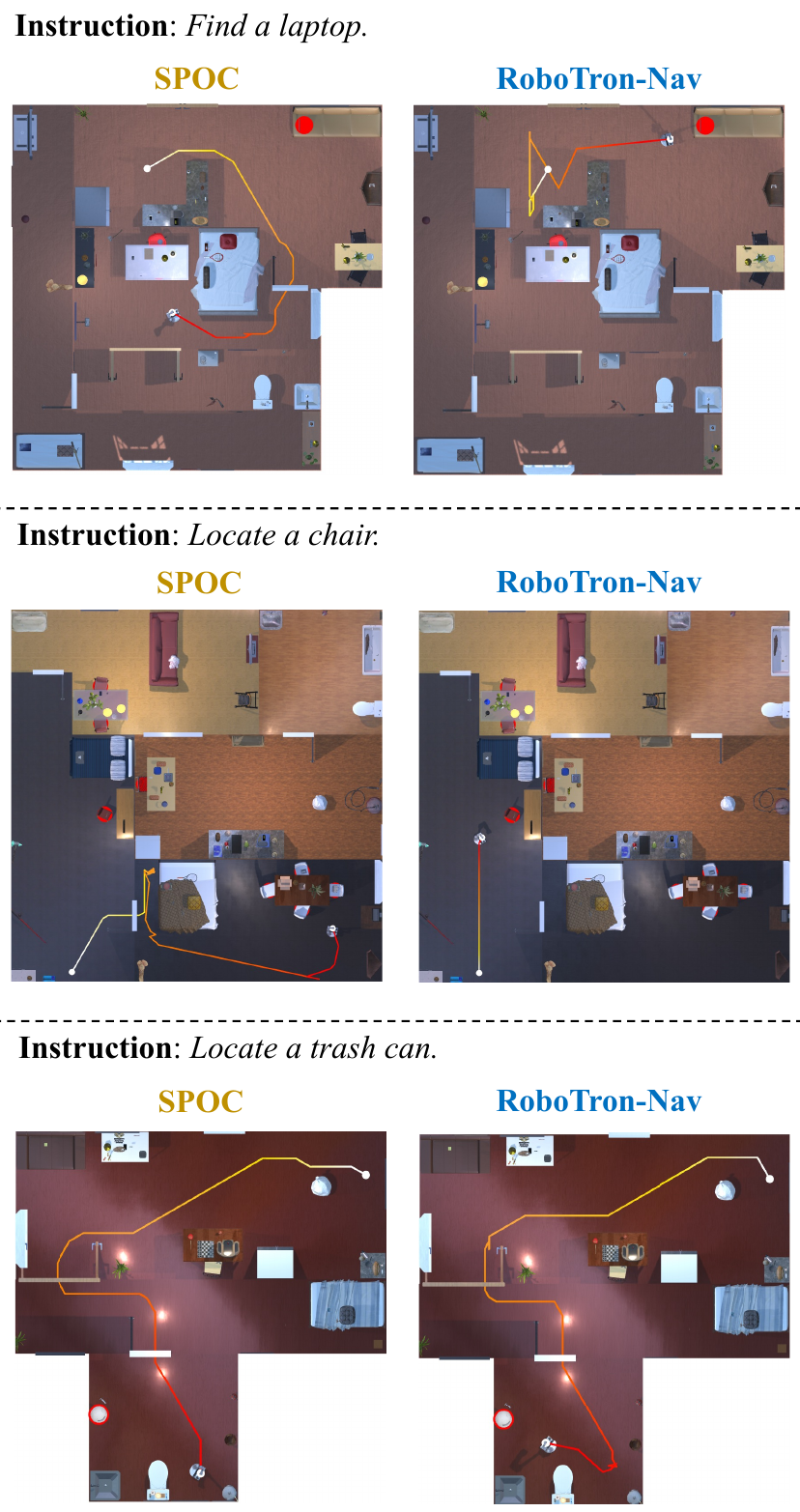}
\caption{Qualitative comparison of trajectories generated by SPOC~\cite{ehsani2024spoc} and RoboTron-Nav in different rooms.}
\label{fig:long}
\end{figure*}

To visualize the effectiveness of our unified framework for embodied navigation, we provide additional qualitative results generated by our method alongside those of SPOC~\cite{ehsani2024spoc}.
As shown in Fig.\ref{fig:short}, when the agent is positioned close to the target, such as within the same room, our RoboTron-Nav is capable of generating the shortest path comparable to SPOC~\cite{ehsani2024spoc}. Both methods understand the environment well, enabling efficient navigation under familiar conditions. 
However, when the agent needs to navigate across greater distances, such as being situated in different rooms from the target, significant differences in their performance begin to emerge. As shown in Fig.~\ref{fig:long}, SPOC~\cite{ehsani2024spoc} struggles by repeating paths, which reduces efficiency and increases the risk of looping or missing optimal routes, lowering its success rate. In contrast, RoboTron-Nav avoids revisiting areas, systematically explores new routes, and adapts to changing environments, making it effective for long-distance navigation and optimizing complex pathways.

